\documentclass[10pt,twocolumn,letterpaper]{article}

\usepackage{wacv}
\makeatletter
\@namedef{ver@everyshi.sty}{}
\makeatother
\usepackage{tikz}
\usepackage{times}
\usepackage{epsfig}
\usepackage{graphicx}
\usepackage{amsmath}
\usepackage{amssymb}
\usepackage{bm}
\usepackage{bbm}
\usepackage{float}

\newenvironment{packed_enum}{
\begin{itemize}
  \setlength{\itemsep}{3pt}
  \setlength{\parskip}{0pt}
  \setlength{\parsep}{0pt}
}{\end{itemize}}

\usepackage{pifont}
\newcommand{\xmark}{\ding{55}}
\usepackage{multirow,makecell, rotating}
\usepackage{booktabs}
\usepackage{tikz}
\usepackage{graphbox}
\usepackage{comment}
\usepackage{epsfig}
\usepackage{bm}
\usepackage{soul} 
\usepackage{color}
\usepackage{verbatim}
\usepackage{etoolbox}
\usepackage{mathtools}
\usepackage{tabularx}
\usepackage{wasysym}

\newcolumntype{Y}{>{\centering\arraybackslash}X}
\DeclarePairedDelimiter\abs{\lvert}{\rvert}%
\makeatletter
\let\oldabs\abs
\def\abs{\@ifstar{\oldabs}{\oldabs*}}
\DeclarePairedDelimiter\norm{\lVert}{\rVert}%
\let\oldnorm\norm
\def\norm{\@ifstar{\oldnorm}{\oldnorm*}}
\makeatother
\usepackage[pagebackref=true,breaklinks=true,letterpaper=true,colorlinks,bookmarks=false]{hyperref}

\def\wacvPaperID{72} 

\wacvalgorithmstrack   

\wacvfinalcopy 

\ifwacvfinal
\usepackage[breaklinks=true,bookmarks=false]{hyperref}
\else
\usepackage[pagebackref=true,breaklinks=true,colorlinks,bookmarks=false]{hyperref}
\fi

\begin{document}

%
\title{Asymmetric Student-Teacher Networks for Industrial Anomaly Detection}
%
%
%

\author{Marco Rudolph\textsuperscript{1}
\and
Tom Wehrbein\textsuperscript{1}
\and
Bodo Rosenhahn\textsuperscript{1}
\and
Bastian Wandt\textsuperscript{2}
\and
\hspace{-2.5mm}\textsuperscript{1}L3S / Leibniz University Hannover, Germany \quad \quad \textsuperscript{2}Linköping University, Sweden\\
{\tt\small rudolph@tnt.uni-hannover.de}
}

\maketitle
\thispagestyle{empty}

\vspace{-10pt}
\begin{abstract}
\vspace{-10pt}
Industrial defect detection is commonly addressed with anomaly detection (AD) methods where no or only incomplete data of potentially occurring defects is available.
This work discovers previously unknown problems of student-teacher approaches for AD and proposes a solution, where two neural networks are trained to produce the same output for the defect-free training examples.
The core assumption of student-teacher networks is that the distance between the outputs of both networks is larger for anomalies since they are absent in training.
However, previous methods suffer from the similarity of student and teacher architecture, such that the distance is undesirably small for anomalies.
For this reason, we propose asymmetric student-teacher networks (AST).
We train a normalizing flow for density estimation as a teacher and a conventional feed-forward network as a student to trigger large distances for anomalies:
The bijectivity of the normalizing flow enforces a divergence of teacher outputs for anomalies compared to normal data.
Outside the training distribution the student cannot imitate this divergence due to its fundamentally different architecture.
Our AST network compensates for wrongly estimated likelihoods by a normalizing flow, which was alternatively used for anomaly detection in previous work.
We show that our method produces state-of-the-art results on the two currently most relevant defect detection datasets MVTec AD and MVTec 3D-AD regarding image-level anomaly detection on RGB and 3D data.

\end{abstract}
\vspace{-10pt}

\section{Introduction}
\label{intro}
\begin{figure}
\vspace*{-10pt}
\centering
  \includegraphics[width=0.45\textwidth]{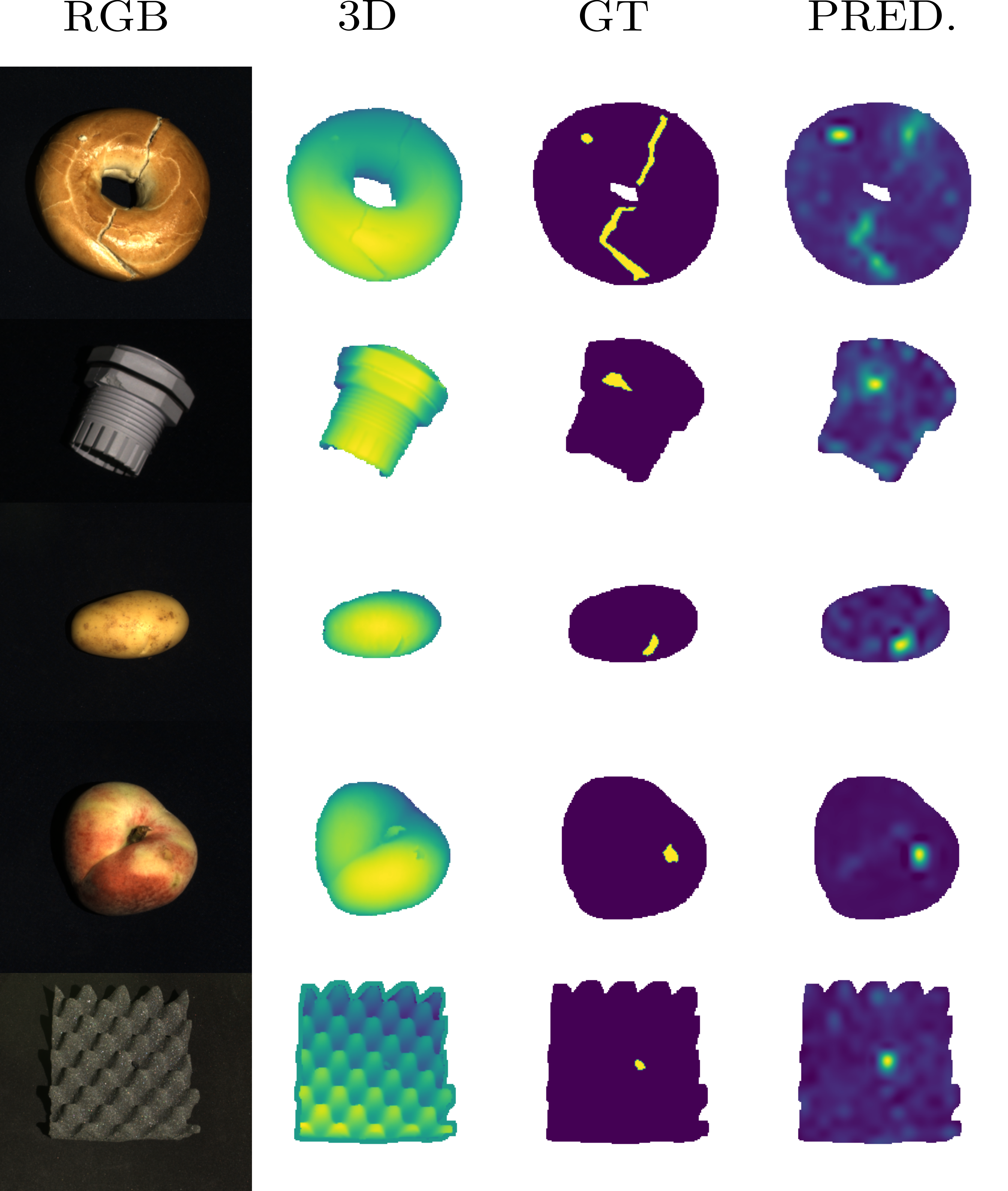}
 \caption{Qualitative results on MVTec 3D-AD~\cite{mvtec3d}. The two left columns show the input, the third the ground truth and the fourth our anomaly detection. Images are masked by foreground extraction. Our method is able to successfully combine RGB and 3D data to detect defects even if only present in one data domain.}
\label{fig:loc}
\vspace*{-10pt}
\end{figure}

To ensure product quality and safety standards in industrial manufacturing processes, products are traditionally inspected by humans, which is expensive and unreliable in practice.
For this reason, image-based methods for automatic inspection have been developed recently using advances in deep learning \cite{ae_ssim, itae, cutpaste, differnet, csflow}.
Since there are no or only very few negative examples, \ie erroneous products, available, especially at the beginning of production, and new errors occur repeatedly during the process, traditional supervised algorithms cannot be applied to this task.
Instead, it is assumed that only data of a \textit{normal} class of defect-free examples is available in training which is termed as semi-supervised \emph{anomaly detection}.
This work and others~\cite{ae_ssim, cflow, patchcore, differnet, csflow} specialize for industrial anomaly detection.
This domain differs in contrast to others that normal examples are similar to each other and to defective products.
In this work, we not only show the effectiveness of our method for common RGB images but also on 3D data and their combination as shown in Figure \ref{fig:loc}.

\begin{figure}
\vspace*{-10pt}
\centering
  \includegraphics[width=0.48\textwidth]{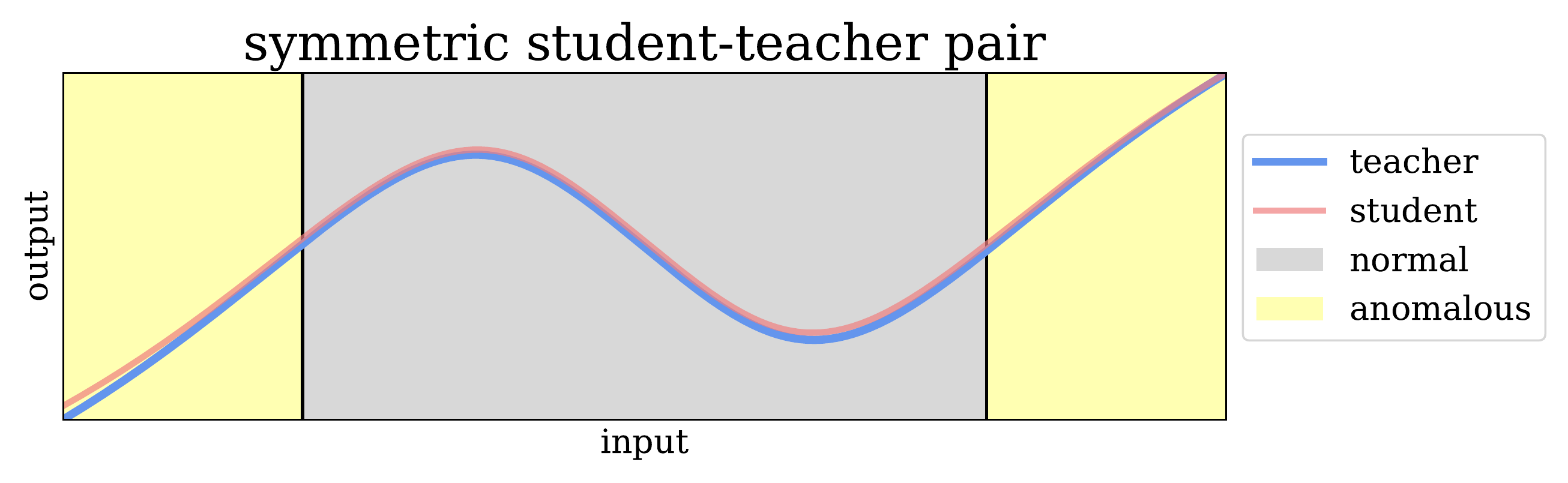} 
  \includegraphics[width=0.48\textwidth]{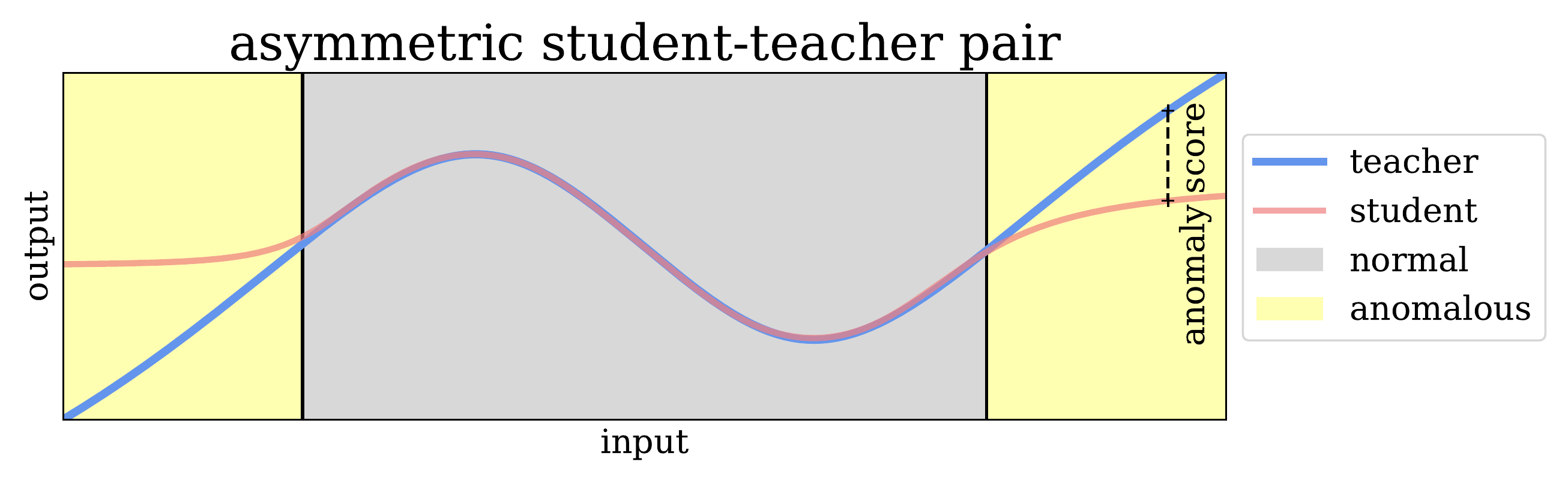} 
 \caption{Toy example with mini-MLPs: The students were optimized to match the outputs in the grey area.
 While the symmetric student-teacher pair (top) generalizes unintentionally and maps anomalous data very similarly, the distance between student and teacher outputs can be used for anomaly detection in the asymmetric student-teacher pair (bottom).}
\label{fig:teaser}
\vspace*{-10pt}
\end{figure}

Several approaches try to solve the problem by so-called \textit{student-teacher networks} \cite{st_bergmann2, st_bergmann1, georgescu2021anomaly, wang2021student, xiao2021unsupervised}.
First, the teacher is trained on a pretext task to learn a semantic embedding.
In a second step, the student is trained to match the output of the teacher.
The motivation is that the student can only match the outputs of the teacher on normal data since it is trained only on normal data.
The distance between the outputs of student and teacher is used as an indicator of an anomaly at test-time.
It is assumed that this distance is larger for defective examples compared to defect-free examples.
However, this is not necessarily the case in previous work, since we discovered that both teacher and student are conventional (i.~e. non-injective) neural networks with similar architecture.
A student with similar architecture tends to undesired generalization, such that it extrapolates similar outputs as the teacher for inputs that are out of the training distribution, which, in turn, gives an undesired low anomaly score.
This effect is shown in Figure \ref{fig:teaser} using an explanatory experiment with one-dimensional data:
If the same neural network with one hidden layer is used for student and teacher, the outputs are still similar for anomalous data in the yellow area of the upper plot.
In contrast, the outputs for anomalies diverge if an MLP with 3 hidden layers is used as the student.

In general, it is not guaranteed that an out-of-distribution input will cause a sufficiently large change in both outputs due to the missing injectivity of common neural networks.
In contrast to normalizing flows, conventional networks have no guarantee to provide out-of-distribution outputs for out-of-distribution inputs.
These problems motivate us to use an asymmetric student-teacher pair (\emph{AST}):
A bijective normalizing flow \cite{nf} acts as a teacher while a conventional sequential model acts as a student.
In this way, the teacher guarantees to be sensitive to changes in the input caused by anomalies.
Furthermore, the usage of different architectures and thus of different sets of learnable functions enforces the effect of distant outputs for out-of-distribution samples.
As a pretext task for the teacher, we optimize to transform the distribution of image features and/or depth maps to a normal distribution via maximum likelihood training which is equivalent to a density estimation~\cite{realnvp}.
This optimization itself is used in previous work~\cite{cflow, differnet, csflow} for anomaly detection by utilizing the likelihoods as an anomaly score:
A low likelihood of being normal should be an indicator of anomalies.
However, Le and Dinh~\cite{le2021perfect} have shown that even perfect density estimators cannot guarantee anomaly detection.
For example, just reparameterizing the data would change the likelihoods of samples.
Furthermore, unstable training leads to misestimated likelihoods.
We show that our student-teacher distance is a better measure for anomaly detection compared to the obtained likelihoods by the teacher.
The advantage to using a normalizing flow itself for anomaly detection is that a possible misestimation in likelihood can be compensated for:
If a low likelihood of being normal is incorrectly assigned to normal data, this output can be predicted by the student, thus still resulting in a small anomaly score.
If a high likelihood of being normal is incorrectly assigned to anomalous data, this output cannot be predicted by the student, again resulting in a high anomaly score.
In this way, we combine the benefits of student-teacher networks and density estimation with normalizing flows.
We further enhance the detection by a positional encoding and by masking the foreground using 3D images.

Our contributions are summarized as follows:
\begin{packed_enum}
    \item Our method avoids the undesired generalization from teacher to student by having highly asymmetric networks as a student-teacher pair.
    \item We improve student-teacher networks by incorporating a bijective normalizing flow as a teacher.
    \item Our AST outperforms the density estimation capability of the teacher by utilizing student-teacher distances.
   \item Code is available on GitHub\footnote{\url{https://github.com/marco-rudolph/ast}}.
\end{packed_enum}

\section{Related Work}

\subsection{Student-Teacher Networks}
Originally, the motivation for having a student network that learns to regress the output of a teacher network was to distill knowledge and save model parameters \cite{hinton2015distilling, mirzadeh2020improved, tian2019contrastive}.
In this case, a student with clearly fewer parameters compared to the teacher almost matches the performance.
Some previous work exploits the student-teacher idea for anomaly detection by using the distance between their outputs:
The larger the distance, the more likely the sample is anomalous.
Bergmann et al.~\cite{st_bergmann1} propose an ensemble of students which are trained to regress the output of a teacher for image patches.
This teacher is either a distilled version of an ImageNet-pre-trained network or trained via metric learning.
The anomaly score is composed of the student uncertainty, measured by the variance of the ensemble, and the regression error.
Wang et al.~\cite{wang2021student} extend the student task by regressing a feature pyramid rather than a single output of a pre-trained network.
Bergmann and Sattlegger \cite{st_bergmann2} adapt the student-teacher concept to point clouds. 
Local geometric descriptors are learned in a self-supervised manner to train the teacher.
Xiao et al.~\cite{xiao2021unsupervised} let teachers learn to classify applied image transformations.
The anomaly score is a weighted sum of the regression error and the class score entropy of an ensemble of students.
By contrast, our method requires only one student and the regression error as the only criterion to detect anomalies. 
All of the existing work is based on identical and conventional (non-injective) networks for student and teacher, which causes undesired generalization of the student as explained in Section~\ref{intro}.

\subsection{Density Estimation}
Anomaly detection can be viewed from a statistical perspective:
By estimating the density of normal samples, anomalies are identified through a low likelihood.
The concept of density estimation for anomaly detection can be simply realized by assuming a multivariate normal distribution.
For example, the Mahalanobis distance of pre-extracted features can be applied as an anomaly score~\cite{padim, rippel} which is equivalent to computing the negative log likelihood of a multivariate Gaussian.
However, this method is very inflexible to the training distributions, since the assumption of a Gaussian distribution is a strong simplification.

To this end, many works try to estimate the density more flexibly with a Normalizing Flow~(NF) \cite{nf_trajectory, cflow, differnet, csflow,  nf_deep, nf_time_series}.
Normalizing Flows are a family of generative models that map bijectively by construction \cite{inn, realnvp,nf,  tomINN} as opposed to conventional neural networks.
This property enables exact density estimation in contrast to other generative models like GANs~\cite{gan} or VAEs~\cite{vae}.
Rudolph et al.~\cite{differnet} make use of this concept by modeling the density of multi-scale feature vectors obtained by pre-trained networks.
Subsequently, they extend this to multi-scale feature maps instead of vectors to avoid information loss caused by averaging~\cite{csflow}.
To handle differently sized feature maps so-called cross-convolutions are integrated.
A similar approach by Gudovskiy et al.~\cite{cflow} computes a density on feature maps with a conditional normalizing flow, where likelihoods are estimated on the level of local positions which act as a condition for the NF.

A common problem of normalizing flows is unstable training, which has a tradeoff on the flexibility of density estimation~\cite{cinn}.
However, even the ground truth density estimation does not provide perfect anomaly detection, since the density strongly depends on the parameterization~\cite{le2021perfect}.

\subsection{Other Approaches}
\hspace{-4.2mm}\textbf{Generative Models}\\
Many approaches try to tackle anomaly detection based on other generative models than normalizing flows as autoencoders~\cite{ae_ssim, itae, memae, sae,dsebm, adae} or GANs~\cite{ganomaly, ADGAN, anogan}.
This is motivated by the inability of these models to generate anomalous data.
Usually, the reconstruction error is used for anomaly scoring.
Since the magnitude of this error depends highly on the size and structure of the anomaly, these methods underperform in the industrial inspection setting.
The disadvantage of these methods is that the synthetic anomalies cannot imitate many real anomalies.\\
\textbf{Anomaly Synthesization}\\
Some work reformulates semi-supervised anomaly detection as a supervised problem by synthetically generating anomalies.
Either parts of training images~\cite{cutpaste, nsa, anoseg} or random images~\cite{draem} are patched into normal images.
Synthetic masks are created to train a supervised segmentation.\\
\textbf{Traditional Approaches}\\
In addition to deep-learning-based approaches, there are also classical approaches for anomaly detection.
The one-class SVM~\cite{ocsvm} is a max-margin method optimizing a function that assigns a higher value to high-density regions than to low-density regions.
Isolation forests~\cite{isoforest} are based on decision trees, where a sample is considered anomalous if it can be separated from the rest of the data by a few constraints.
Local Outlier Factor~\cite{lof} compares the density of a point with that of its neighbors.
A comparatively low density of a point identifies anomalies.
Traditional approaches usually fail in visual anomaly detection due to the high dimensionality and complexity of the data. 
This can be circumvented by combining them with other techniques:
For example, the distance to the nearest neighbor, as first proposed by Amer and Goldstein~\cite{amer2012nearest}, is used as an anomaly score after features are extracted by a pre-trained network~\cite{nazare,patchcore}.
Alternatively point cloud features~\cite{fpfh} or density-based clustering~\cite{DocBra2015, DocBra2016} can be used to characterize a points neighborhood and label it accordingly.
However, the runtime is linearly related to the dataset size.

\section{Method}
\label{overview}
Our goal is to train two models, a student model $f_s$ and a teacher model $f_t$, such that the student learns to regress the teacher outputs on defect-free image data only.
The training process is divided into two phases:
First, the teacher model is optimized to transform the training distribution $p_X$ to a normal distribution $\mathcal{N}(0,\,I)$ bijectively with a normalizing flow.
Second, the student is optimized to match the teacher outputs by minimizing the distance between $f_s(x)$ and $f_t(x)$ of training samples $x \in X$.
We apply the distance for anomaly scoring at test-time, which is further described in Section~\ref{student}.

We follow~\cite{st_bergmann1, cflow, csflow} and use extracted features obtained by a pre-trained network on ImageNet~\cite{imagenet} instead of RGB images as direct input for our models.
Such networks have been shown to be universal feature extractors whose outputs carry relevant semantics for industrial anomaly detection.

In addition to RGB data, our approach is easily extendable to multimodal inputs including 3D data.
If 3D data is available, we concatenate depth maps to these features along the channels.
Since the feature maps are reduced in height and width compared to the depth map resolution by a factor $d$, we apply pixel-unshuffling~\cite{pixelunshuffle} by grouping a depth image patch of $d \times d$ pixels as one pixel with $d^2$ channels to match the dimensions of the feature maps.

Any 3D data that may be present is used to extract the foreground.
This is straightforward and reasonable whenever the background is static or planar, which is the case for almost all real applications.
Pixels that are in the background are ignored when optimizing the teacher and student by masking the distance and negative log likelihood loss, which are introduced in Sections \ref{teacher} and \ref{student}.
If not 3D data is available, the whole image is considered as foreground.
Details of the foreground extraction are given in Section~\ref{preprocessing}.

\begin{figure}
\centering
  \includegraphics[width=0.47\textwidth]{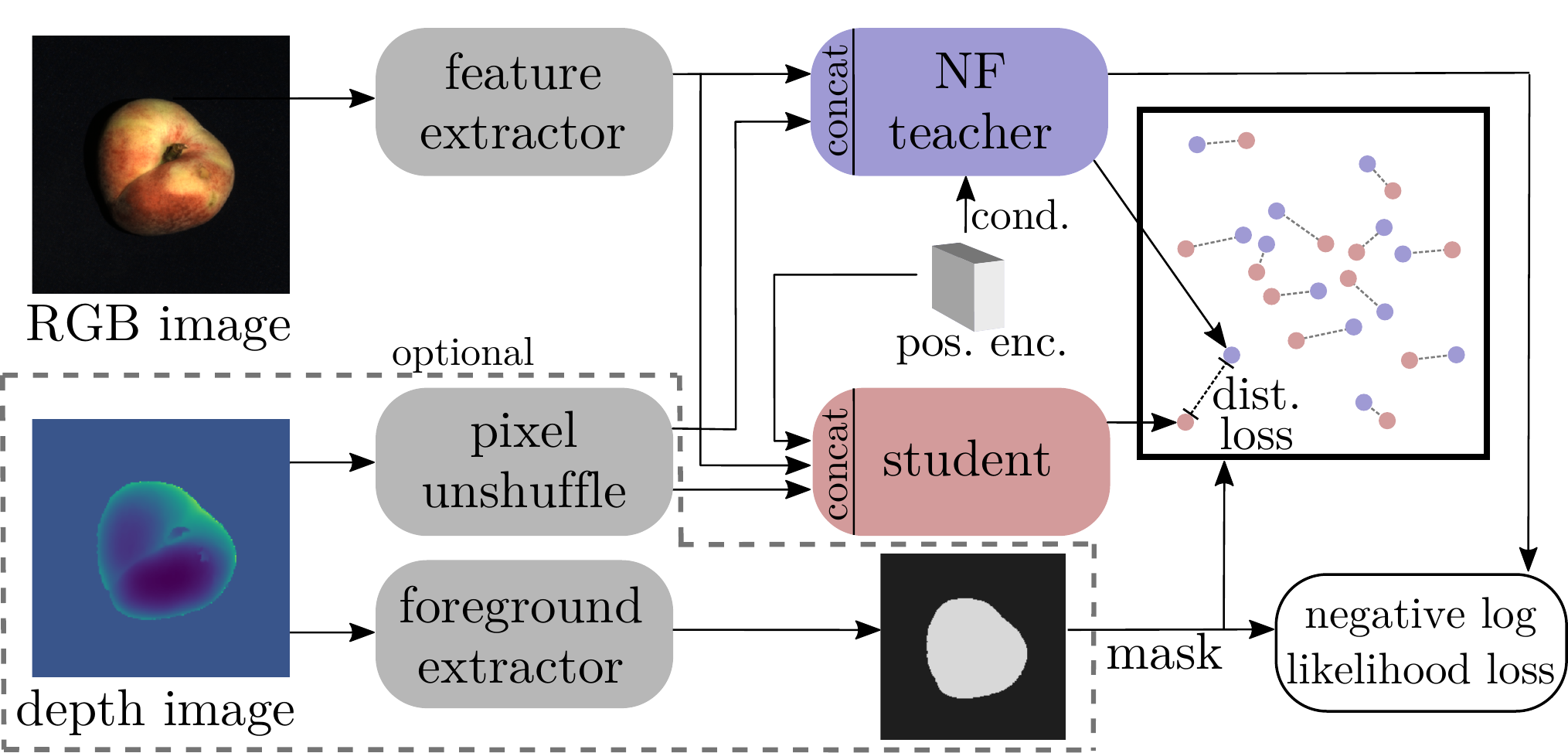} 
 \caption{
 Overview of our pipeline: Teacher and student receive image features and/or depth maps as input which is conditioned by a positional encoding.
 First, the teacher represented by a normalizing flow is optimized to reduce the negative log likelihood loss that may be masked by a foreground map from 3D.
 Second, the student is trained to match the teacher outputs by minimizing the (masked) distance between them.
 }
\label{fig:overview}
\vspace*{-10pt}
\end{figure}

Similar to~\cite{cflow}, we use a sinusoidal positional encoding~\cite{posenc} for the spatial dimensions of the input maps as a condition for the normalizing flow $f_t$.
In this way, the occurrence of a feature is related to its position to detect anomalies such as misplaced objects.
An overview of our pipeline is given in Figure~\ref{fig:overview}.

\begin{figure}
\centering
  \includegraphics[width=0.47\textwidth]{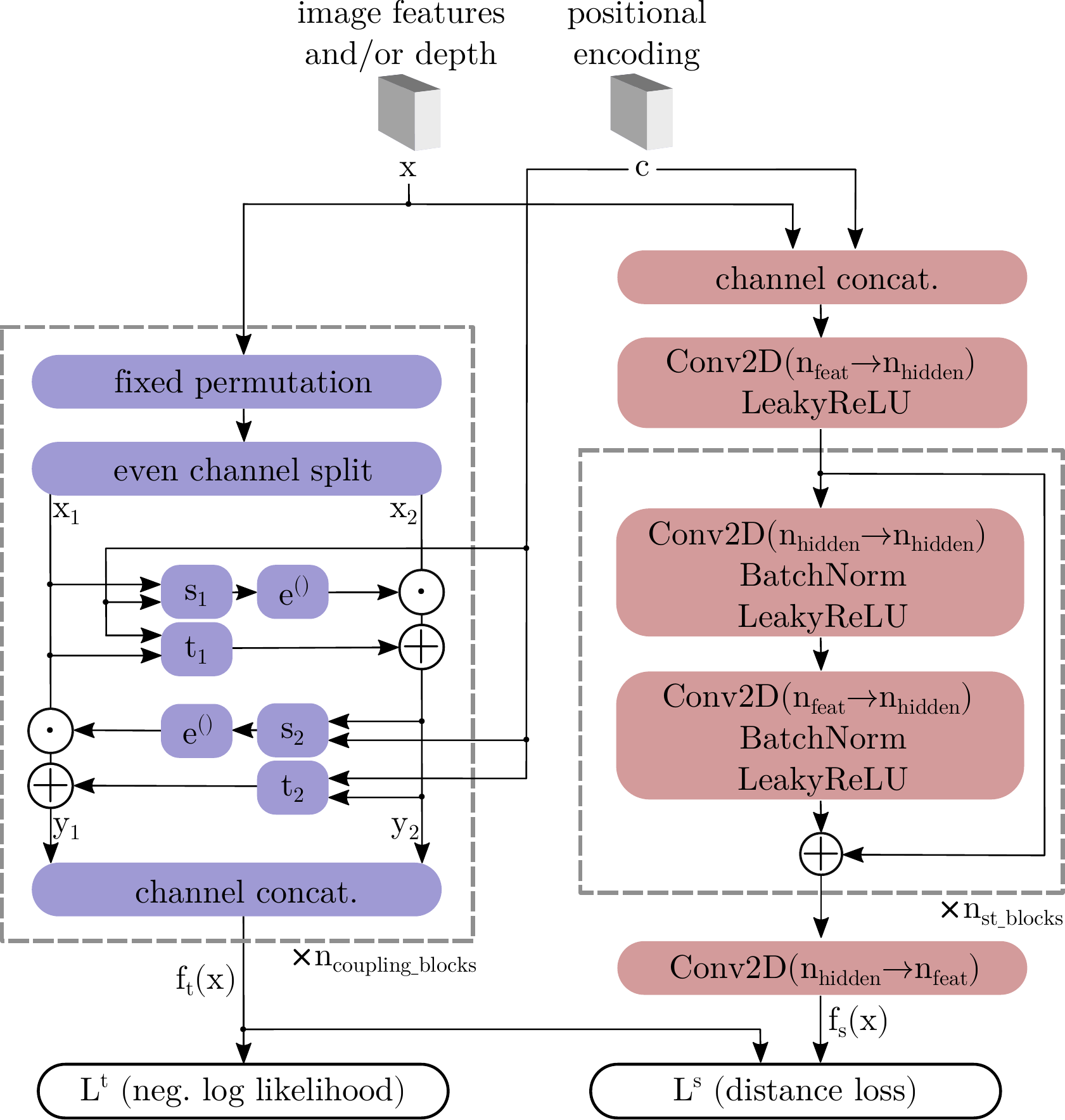} 
 \caption{Model architecture of teacher (left side) and student (right side). While the teacher is a Real-NVP-based~\cite{realnvp} conditional normalizing flow~\cite{cinn}, the student is a conventional convolutional neural network.}
\label{fig:architecture}
\end{figure}

\subsection{Teacher}
\label{teacher}
Similar to \cite{cflow, differnet, csflow}, we train a normalizing flow based on Real-NVP~\cite{realnvp} to transform the training distribution to a normal distribution $\mathcal{N}(0,\,I)$.
In contrast to previous work, we do not use the outputs to compute likelihoods and thereby obtain anomaly scores directly.
Instead, we interpret this training as a pretext task to create targets for our student network.

The normalizing flow consists of multiple subsequent affine coupling blocks.
Let the input $x \in \mathbb{R}^{w\times h \times n_{\mathrm{feat}}}$ be feature maps with $n_{\mathrm{feat}}$ features of size $w\times h$.
Within these blocks, the channels of the input $x$ are split evenly along the channels into the parts $x_1$ and $x_2$ after randomly choosing a permutation that remains fixed.
\interfootnotelinepenalty=10000
These parts are each concatenated with a positional encoding $c$ as a static condition.
Both are used to compute scaling and shift parameters for an affine transformation of the counterpart by having subnetworks $s_i$ and $t_i$ for each part:
\begin{equation}
 \begin{aligned}
y_2 = x_2 \odot e^{s_1([x_1, c])} + t_1([x_1, c])  \\
y_1 = x_1 \odot e^{s_2([x_2, c])} + t_2([x_2, c]),
\end{aligned}
\label{aff}
\end{equation}
where $\odot$ is the element-wise product and $[\cdot , \cdot]$ denotes concatenation. 
The output of one coupling block is the concatenation of $y_1$ and $y_2$ along the channels.
Note that the number of dimensions of input and output does not change due to invertibility.

To stabilize training, we apply alpha-clamping of scalar coefficients as in~\cite{cinn} and the gamma-trick as in \cite{csflow}.
Using the change-of-variable formula with $z$ as our final output
\begin{equation}
\label{eqn:change_of_variables}
    p_X(x) = p_Z(z) \abs{
    \det{
    \frac{\partial z}
        {\partial x}
    }}\quad ,
\end{equation}
we minimize the negative log likelihood with $p_Z$ as the normal distribution $\mathcal{N}(0,\,I)$ by optimizing the mean of
\begin{equation}
 \begin{aligned}
    \mathcal{L}_{ij}^t = -\log{p_X(x_{ij})} =  \frac{\norm{z_{ij}}_2^2}{2}    - \log{\abs{
    \det{
    \frac{\partial z_{ij}}
        {\partial x_{ij}}
    }}}
 \end{aligned}
 \label{formula:loglikelihood}
\end{equation}
over all (foreground) pixels at pixel position $(i,j)$.

\begin{table}
\small
\begin{center}
\footnotesize
    \begin{tabular}{|l|c|c|}
    \hline
    Dataset & MVTec AD & MVTec 3D-AD\\ 
    Alias & (MVT2D) & (MVT3D) \\ \hline 
    RGB images & \checkmark & \checkmark\\
    3D scans & $\times$ & \checkmark \\
    \#categories & 15 & 10 \\
    image side length & 700-1024 & 400-800 \\
    \#train samples per cat. & 60-320 & 210-300\\
    \#test samples per cat. & 42-160 & 100-159\\
    \#defect types per cat. & 1-7 & 3-5\\
    \hline
    \end{tabular}
\end{center}
\vspace{-0.25cm}
\caption{Overview of the used datasets.}
    \vspace{-1.0em}
    \label{table:datasets}
\end{table}

\begin{table*}

\begin{center}\resizebox{0.9\linewidth}{!}{
\hskip-0.5cm
\footnotesize

\begin{tabular}{c|l|cccccccccccc|}
\cline{2-14}
& Category & ARNet & DR\AE M & GAN & Rippel & PatchCore & DifferNet & PaDiM & CFlow & CS-Flow & Uninf. & STFPM* & \textbf{AST}\\ 
 & &\cite{itae}& \cite{draem} &\cite{ganomaly} &  \cite{rippel} & \cite{patchcore}& \cite{differnet} & \cite{padim} & \cite{cflow} & \cite{csflow} & Stud. \cite{st_bergmann1} & \cite{wang2021student} & (ours)\\
\cline{2-14}
& Grid       & 88.3 & 99.9 & 70.8   & 93.7 & 98.2 & 84.0 & - & 99.6 & 99.0 & 98.1 & \textbf{100} & 99.1 $\pm$ 0.2\\
& Leather    & 86.2 & 100 & 84.2  &\textbf{100} & \textbf{100} & 97.1 & - & \textbf{100} &  99.9 & 94.7 &  \textbf{100} & \textbf{100} $\pm$ 0.0\\
& Tile       & 73.5 & 99.6 & 79.4 & \textbf{100} & 98.7 & 99.4 & - & 99.9 & \textbf{100} &  94.7 & 95.5 & \textbf{100} $\pm$ 0.0\\
& Carpet     & 70.6 & 97.8 & 69.9  & 99.6 & 98.7 & 92.9 & - & 98.7 & \textbf{100} & 99.9 & 98.9 & 97.5 $\pm$ 0.4\\
\rotatebox[origin=c]{90}{\parbox[c]{0cm}{Textures}}& Wood       & 92.3 & 99.1 & 83.4  & 99.2 & 98.8 & 99.8 & - & 99.1 & \textbf{100} & 99.1 & 99.2 & \textbf{100} $\pm$ 0.0\\
\cline{2-14}
& Avg. Text. & 82.2 & 99.3 & 77.5 & 98.5 & 98.3 & 94.6 & 99.0 & 99.5&\textbf{99.8} & 97.3 & 98.7 & 99.3 $\pm$ 0.08\\
\cline{2-14}
& Bottle     & 94.1 & 99.2 & 89.2 & 99.0 & \textbf{100} & 99.0  & - & \textbf{100} &  99.8 & 99.0 & \textbf{100} & \textbf{100} $\pm$ 0.0\\
& Capsule    & 68.1 & 98.5 & 73.2 & 96.3 & 98.1 & 86.9 & - & 97.7 & 97.1 & 92.5 & 88.0 & \textbf{99.7} $\pm$ 0.1\\
& Pill       & 78.6 & 98.9 & 74.3 & 91.4 & 96.6 & 88.8 & - &  96.8 & 98.6 & 92.2 & 93.8 & \textbf{99.1} $\pm$ 0.1\\
& Transistor & 84.3 & 93.1 & 79.2 & 98.2 & \textbf{100} & 91.1 & - & 95.2 & 99.3 & 79.4 & 93.7 & 99.3 $\pm$ 0.1\\
& Zipper     & 87.6 & \textbf{100} & 74.5 & 98.8 & 99.4 & 95.1 & - & 98.5 & 99.7 &  94.4 & 93.6  & 99.1 $\pm$ 0.1\\
& Cable      & 83.2 & 91.8 & 75.7 & 99.1 & \textbf{99.5} & 95.9 & - & 97.6 & 99.1 & 78.7 & 92.3 & 98.5 $\pm$ 0.2 \\
\rotatebox[origin=c]{90}{\parbox[c]{0cm}{Objects}}& Hazelnut   & 85.5 & \textbf{100} & 78.5 & \textbf{100} & \textbf{100} & 99.3 & - & \textbf{100} & 99.6 & 99.1 & \textbf{100} & \textbf{100} $\pm$ 0.0 \\
& Metal Nut  & 66.7 & 98.7 & 70.0 & 97.4 & \textbf{100} & 96.1 & - & 99.3 & 99.1 &  89.1 & \textbf{100} & 98.5 $\pm$ 0.2\\
& Screw      & \textbf{100} & 93.9 & 74.6  & 94.5 & 98.1 & 96.3 & - & 91.9 & 97.6 & 86.0 & 88.2 & 99.7 $\pm$ 0.1\\
& Toothbrush & \textbf{100} & \textbf{100} & 65.3 & 94.1 & \textbf{100} & 98.6 & - &  99.7 & 91.9 & \textbf{100} & 87.8 & 96.6 $\pm$ 0.1\\
\cline{2-14}
& Avg. Obj.  & 84.8 & 97.4 & 75.5 & 96.9 & \textbf{99.2} & 94.7 & 97.2& 97.7 & 98.2 & 91.0 & 93.7 &  99.1 $\pm$ 0.03\\
\cline{2-14}
& \textbf{Average} & 83.9 & 98.0 & 76.2 & 97.5 & 99.1 & 94.7 & 97.9 & 98.3 & 98.7 & 93.2 & 95.4 & \textbf{99.2} $\pm$ 0.04\\
\cline{2-14}
\end{tabular}
}
\end{center}
\caption{AUROC in \% for detecting defects of all categories of MVT2D \cite{mvtec} on image-level grouped into textures and objects. We report the mean and standard deviation over 5 runs for our method. Best results are in bold.
Beside the average value, detailed results of PaDiM~\cite{padim} were not provided by the authors. The numbers of STFPM*~\cite{wang2021student} were obtained by a reimplementation.
}
\label{table:mvtec}
\end{table*}

\subsection{Student}
\label{student}
As opposed to the teacher, the student is a conventional feed-forward network that does not map injectively or surjectively.
We propose a simple fully convolutional network with residual blocks which is shown in Figure~\ref{fig:architecture}.
Each residual block consists of two sequences of $3 \times 3$ convolutional layers, batch normalization~\cite{batchnorm} and leaky ReLU activations.
We add convolutions as the first and last layer to increase and decrease the feature dimensions.

Similarly to the teacher, the student takes image features as input which are concatenated with 3D data if available.
In addition, the positional encoding $c$ is concatenated.
The output dimensions match the teacher to enable pixel-wise distance computation.
We minimize the squared $\ell_2$-distance between student outputs $f_s(x)$ and the teacher outputs $f_t(x)$ on training samples $x \in X$, given the training set $X$, at a pixel position $(i, j)$ of the output:

\begin{equation}
\mathcal{L}^s_{ij} = \norm{f_s(x)_{ij} - f_t(x)_{ij}}^2_2 .
\label{eqn: st_loss}
\end{equation}
Averaging $\mathcal{L}^s_{ij}$ over all (foreground) pixels gives us the final loss.
The distance $\mathcal{L}^s$ is also used in testing to obtain an anomaly score on image level:
Ignoring the anomaly scores of background pixels, we aggregate the pixel distances of one sample by computing either the maximum or the mean over the pixels.

\section{Experiments}
\subsection{Datasets}
\label{datasets}
To demonstrate the benefits of our method on a wide range of industrial inspection scenarios, we evaluate with a diverse set of 25 scenarios in total, including natural objects, industrial components and textures in 2D and 3D.
Table~\ref{table:datasets} shows an overview of the used benchmark datasets MVTec AD~\cite{mvtec} and MVTec 3D-AD~\cite{mvtec3d}.
For both datasets, the training set only contains defect-free data and the test set contains defect-free and defective examples.
In addition to image-level labels, the datasets also provide pixel-level annotations about defective regions which we use to evaluate the segmentation of defects.

MVTec AD, which will be called \textit{MVT2D} in the following, is a high-resolution 2D RGB image dataset containing 10 object and 5 texture categories.
The total of 73 defect types in the test set appear, for example, in the form of displacements, cracks or scratches in various sizes and shapes.
The images have a side length of 700 to 1024 pixels.

MVTec 3D-AD, to which we refer to as \textit{MVT3D}, is a very recent 3D dataset containing 2D RGB images paired with 3D scans for 10 categories.
These categories include deformable and non-deformable objects, partially with natural variations (e.g.\ peach and carrot).
In addition to the defect types in MVT2D there are also defects that are only recognizable from the depth map, such as indentations.
On the other hand, there are anomalies such as discoloration that can only be perceived from the RGB data.
The RGB images have a resolution of 400 to 800 pixels per side, paired with rasterized 3D point clouds at the same resolution.

\subsection{Implementation Details}

\label{impdetails}
\subsubsection{Image Preprocessing}
\label{preprocessing}
Following \cite{padim, csflow}, we use the layer 36 output of EfficientNet-B5~\cite{efficientnet} pre-trained on ImageNet~\cite{imagenet} as a feature extractor.
This feature extractor is not trained during training of the student and teacher networks.
The images are resized to a resolution of $768\times768$ pixels resulting in feature maps of size $24\times24$ with 304 channels.
\subsubsection{3D Preprocessing}
We discard the $x$ and $y$ coordinates due to the low informative content and use only the depth component $z$ in centimeters.
Missing depth values are repeatedly filled by using the average value of valid pixels from an 8-connected neighborhood for 3 iterations.
We model the background as a 2D plane by interpolating the depth of the 4 corner pixels.
A pixel is assumed as foreground if its depth is further than $7mm$ distant from the background plane.
As an input to our models, we first resize the masks to $192\times192$ pixels via bilinear downsampling and then perform pixel-unshuffling~\cite{pixelunshuffle} with $d=8$ as described in Section~\ref{overview} to match the feature map resolution.
In order to detect anomalies at the edge of the object and fill holes of missing values, the foreground mask is dilated using a square structural element of size 8.
We subtract the mean foreground depth from each depth map and set its background pixels to 0.
The binary foreground mask $M$ with ones as foreground and zeros as background is downsampled to feature map resolution to mask the loss for student and teacher.
This is done by a bilinear interpolation $f_\downarrow$ followed by a binarization where all entries greater than zero are assumed as foreground to mask the loss at position~$(i, j)$:
\begin{equation}
    \mathcal{L}^{\mathrm{masked}}_{ij} = 
      \begin{cases}
        \mathcal{L}_{ij} & \text{if }\quad f_\downarrow(M)_{ij} > 0 \\
        0 & else
      \end{cases}
      .
\label{eq:mask}
\end{equation}

\subsubsection{Teacher}
For the normalizing flow architecture of the teacher, we use 4 coupling blocks which are conditioned on a positional encoding with 32 channels.
Each pair of internal subnetworks $s_i$ and $t_i$ is designed as one shallow convolutional network $r_i$ with one hidden layer whose output is split into the scale and shift components.
Inside $r_i$ we use ReLU-Activations and a hidden channel size of 1024 for MVT2D and 64 for MVT3D.
We choose the alpha-clamping parameter $\alpha=3$ for MVT2D and $\alpha=1.9$ for MVT3D.
The teacher networks are trained for 240 epochs for MVT2D and 72 epochs for MVT3D, respectively, with the Adam optimizer~\cite{adam}, using author-given momentum parameters $\beta_1=0.9$ and $\beta_2=0.999$, a learning rate of $2 \cdot 10^{-4}$ and a weight decay of $10^{-5}$.

\subsubsection{Student}
For the student networks, we use $n_{\mathrm{st\_blocks}}=4$ residual convolutional blocks as described in Section \ref{student}.
The Leaky-ReLU-activations use a slope of 0.2 for negative values.
We choose a hidden channel size of $n_{hidden}=1024$ for the residual block.
Likewise, we take over the number of epochs and optimizer parameters from the teacher.
The scores at feature map resolution are aggregated for evaluation at image level by the maximum distance if a foreground mask is available, and the average distance otherwise (RGB only).

\subsection{Evaluation Metrics}
\label{metrics}
As common for anomaly detection, we evaluate the performance of our method on image-level by calculating the area under receiver operating characteristics (AUROC).
The ROC measures the true positive rate dependent on the false positive rate for varying thresholds of the anomaly score.
Thus, it is independent of the choice of a threshold and invariant to the class balance in the test set.
For measuring the segmentation of anomalies at pixel-level, we compute the AUROC on pixel level given the ground truth masks in the datasets.

\begin{table*}[t]
\begin{center}
\footnotesize

\resizebox{0.98\linewidth}{!}{
\begin{tabular}{@{\hskip5pt}l@{\hskip5pt}l|cccccccccc|c} 
    \toprule
\footnotesize
    & Method & Bagel & \begin{tabular}[c]{@{}c@{}}Cable\\ Gland\end{tabular} & Carrot & Cookie & Dowel & Foam & Peach & Potato & Rope & Tire & Mean \\ 
    \midrule
    \multirow{8}{*}{\rotatebox[origin=c]{90}{3D}}
\footnotesize
    & Voxel GAN \cite{mvtec3d}& 38.3 & 62.3  & 47.4 & 63.9 & 56.4 & 40.9 & 61.7 & 42.7 & 66.3 & 57.7 & 53.7 \\
    & Voxel AE \cite{mvtec3d}& 69.3 & 42.5  & 51.5 & 79.0 & 49.4 & 55.8 & 53.7 & 48.4 & 63.9 & 58.3 & 57.1 \\
    & Voxel VM \cite{mvtec3d}& 75.0 & \textbf{74.7}  & 61.3 & 73.8 & 82.3 & 69.3 & 67.9 & 65.2 & 60.9 & \textbf{69.0} & 69.9 \\ 
    & Depth GAN \cite{mvtec3d}& 53.0 & 37.6  & 60.7 & 60.3 & 49.7 & 48.4 & 59.5 & 48.9 & 53.6 & 52.1 & 52.3 \\
    
    & Depth AE \cite{mvtec3d}& 46.8 & 73.1  & 49.7 & 67.3 & 53.4 & 41.7 & 48.5 & 54.9 & 56.4 & 54.6 & 54.6 \\
    
    & Depth VM \cite{mvtec3d}&  51.0 & 54.2  & 46.9 & 57.6 & 60.9 & 69.9 & 45.0 & 41.9 & 66.8 & 52.0 & 54.6 \\
    
    & 1-NN (FPFH) \cite{fpfh}& 82.5 & 55.1 & 95.2 & 79.7 & \textbf{88.3} & 58.2 & 75.8 & 88.9 & 92.9 & 65.3 & 78.2 \\
    
    & 3D-ST$_{128}$  \cite{st_bergmann2}\phone& 86.2 & 48.4 & 83.2 & 89.4 & 84.8 & 66.3 & 76.3 & 68.7 & 95.8 & 48.6 & 74.8 \\
    & \textbf{AST (ours)} & \textbf{88.1} $\pm$ 2.0 & 57.6 $\pm$ 6.9 & \textbf{96.5} $\pm$ 1.0 & \textbf{95.7} $\pm$ 0.6 & 67.9 $\pm$ 1.1 & \textbf{79.7} $\pm$ 1.2 & \textbf{99.0} $\pm$ 0.9 & \textbf{91.5} $\pm$ 2.1 &  \textbf{95.6} $\pm$ 0.7 & 61.1 $\pm$ 3.4 & \textbf{83.3} $\pm$ 0.8 \\

    \midrule
    \multirow{5}{*}{\rotatebox[origin=c]{90}{RGB}}
    & PatchCore \cite{patchcore} & 87.6 & 88.0 & 79.1 & 68.2 & 91.2 & 70.1 & 69.5 & 61.8 & 84.1 & 70.2 & 77.0 \\
    & DifferNet \cite{differnet}\phone & 85.9 & 70.3 & \textbf{64.3} & 43.5 & 79.7 & 79.0 & 78.7 & 64.3 & 71.5 & 59.0 & 69.6 \\
    & PADiM \cite{padim}* & \textbf{97.5} & 77.5 & 69.8 & 58.2 & 95.9 & 66.3 & 85.8 & 53.5 & 83.2 & 76.0 & 76.4 \\
    & CS-Flow \cite{csflow}\phone & 94.1 & \textbf{93.0} & 82.7 & 79.5 & \textbf{99.0} & 88.6 & 73.1 & 47.1 & 98.6 & 74.5 & 83.0 \\
    & STFPM \cite{wang2021student}* & 93.0 & 84.7 & \textbf{89.0} & 57.5 & 94.7 & 76.6 & 71.0 & 59.8 & 96.5 &  70.1 & 79.3 \\
    & \textbf{AST (ours)} & 94.7 $\pm$ 0.7 & 92.8  $\pm$ 1.2 & 85.1 $\pm$ 1.2 & \textbf{82.5} $\pm$ 0.8 & 98.1  $\pm$ 0.4 & \textbf{95.1} $\pm$ 0.6 & \textbf{89.5}  $\pm$ 1.1 & 61.3  $\pm$ 2.4 &  \textbf{99.2} $\pm$ 0.2 & \textbf{82.1} $\pm$ 0.9 & \textbf{88.0} $\pm$ 0.6 \\
    \midrule
    \multirow{8}{*}{\rotatebox[origin=c]{90}{3D + RGB}}
    & Voxel GAN \cite{mvtec3d}& 68.0 & 32.4  & 56.5 & 39.9 & 49.7 & 48.2 & 56.6 & 57.9 & 60.1 & 48.2 & 51.7 \\
    & Voxel AE \cite{mvtec3d}& 51.0 & 54.0  & 38.4 & 69.3 & 44.6 & 63.2 & 55.0 & 49.4 & 72.1 & 41.3 & 53.8 \\
    & Voxel VM \cite{mvtec3d}& 55.3 & 77.2  & 48.4 & 70.1 & 75.1 & 57.8 & 48.0 & 46.6 & 68.9 & 61.1 & 60.9 \\
    & Depth GAN \cite{mvtec3d}& 53.8 & 37.2  & 58.0 & 60.3 & 43.0 & 53.4 & 64.2 & 60.1 & 44.3 & 57.7 & 53.2 \\
    & Depth AE \cite{mvtec3d}& 64.8 & 50.2  & 65.0 & 48.8 & 80.5 & 52.2 & 71.2 & 52.9 & 54.0 & 55.2 & 59.5 \\
    & Depth VM \cite{mvtec3d}& 51.3 & 55.1  & 47.7 & 58.1 & 61.7 & 71.6 & 45.0 & 42.1 & 59.8 & 62.3 & 55.5 \\
    & PatchCore+FPFH \cite{fpfh} & 91.8 & 74.8 & 96.7 & 88.3 & \textbf{93.2} & 58.2 & 89.6 & 91.2 & 92.1 & \textbf{88.6} & 86.5 \\
    
    & \textbf{AST (ours)} & \textbf{98.3} $\pm$ 0.4 & \textbf{87.3} $\pm$3.3& \textbf{97.6} $\pm$ 0.5& \textbf{97.1} $\pm$ 0.3& \textbf{93.2}$\pm$2.1 & \textbf{88.5} $\pm$ 1.4 & \textbf{97.4}$\pm$ 1.4 & \textbf{98.1} $\pm$ 1.2 &  \textbf{100} $\pm$ 0.0 & 79.7 $\pm$ 1.0 & \textbf{93.7}$\pm$ 0.2 \\
  \bottomrule
    \end{tabular}
}
\end{center}
\caption{AUROC in \% for detecting defects of all categories of MVT3D \cite{mvtec3d} on image-level for 3D data, RGB data and the combination of both. We report the mean and standard deviation over 5 runs for our method. Best results per data domain are in bold. Numbers of listed methods followed by a \phone\ are non-published results obtained by the corresponding authors on request. A * indicates that we used a reimplementation. The numbers from PatchCore are taken from \cite{fpfh}.}
\label{table:mvt3d}
\vspace{-0.25cm}
\end{table*}

\begin{table}
\small
\begin{center}
\footnotesize
    \begin{tabular}{l|c|c}
Method &  MVT2D & MVT3D (RGB+3D) \\
    \hline
    AE-SSIM \cite{ae_ssim}& 87.0 & -\\
    PatchCore \cite{patchcore}& \textbf{98.4} & -\\
    PatchCore+FPFH \cite{fpfh}& - & \textbf{99.2}\\
    \textbf{AST (ours)} & 95.0 $\pm$ 0.03 & 97.6 $\pm$ 0.02
    \end{tabular}
\end{center}
    \vspace{-1mm}
\caption{Anomaly segmentation results measured by the mean pixel-AUROC over all classes and its standard deviation over 5 runs. Despite image-level detection is the focus of this work, our method is able to localize defects for practical purposes with an AUROC of 95\% or 97.6\%.}
    \vspace{-5mm}
    \label{tab:seg}
\end{table}

\subsection{Results}
\subsubsection{Detection}
Table \ref{table:mvtec} shows the AUROC of our method and previous work for detecting anomalies on the 15 classes of MVT2D as well as the averages for textures, objects and all classes.
We set a new state-of-the-art performance on the mean detection AUROC over all classes, improving it slightly to 99.2\%.
This is mainly due to the good performance on the more challenging objects, where we outperform previous work by a comparatively large margin of 0.9\%, except for PatchCore~\cite{patchcore}.
The detection of anomalies on textures, which CS-Flow~\cite{csflow} has already almost solved with a mean AUROC of 99.8\%, still works very reliably at 99.3\%.
Especially compared to the two student-teacher approaches \cite{st_bergmann1, wang2021student}, a significant improvement of 6\% and 3.6\% respectively is archieved.
Moreover, our student-teacher distances show to be a better indicator of anomalies compared to the likelihoods of current state-of-the-art density estimators \cite{cflow, csflow} which, like our teacher, are based on normalizing flows.

Even though MVT2D has established itself as a standard benchmark in the past, this dataset (especially the textures) is easily solvable for recent methods, and differences are mainly in the sub-percent range, which is only a minor difference in terms of the comparatively small size of the dataset.
In the following, we focus on the newer, more challenging MVT3D dataset where the normal data shows more variance and anomalies only partly occur in one of the two data modalities, RGB and 3D.

The results for individual classes of MVT3D grouped by data modality are given in Table \ref{table:mvt3d}. 
We are able to outperform all previous methods for all data modalities regarding the average of all classes by a large margin of 5.1\% for 3D, 5\% for RGB and 7.2\% for the combination.
Facing the individual classes and data domains, we set a new state-of-the-art in 21 of 30 cases.
Note that this data set is much more challenging when comparing the best results from previous work (99.1\% for MVT2D vs. 86.5\% AUROC for MVT3D).
Nevertheless, we detect defects in 7 out of 10 cases for RGB+3D at an AUROC of at least 93\%, which demonstrates the robustness of our method.
In contrast, the nearest-neighbor approach PatchCore~\cite{patchcore}, which provides comparable performance to us on MVT2D, struggles with the increased demands of the dataset and is outperformed by 11\% on RGB.
The same applies for the 3D extension~\cite{fpfh} using FPFH~\cite{fpfh_orig} despite using a foreground mask as well.
Figure \ref{fig:loc} shows qualitative results for the RGB+3D case given both inputs and ground truth annotations.
More examples can be found in the supplemental material.
Despite the low resolution, the regions of the anomaly can still be localized well for practical purposes.
Table \ref{tab:seg} reports the pixel-AUROC of our method and previous work.

\begin{figure}
\centering
  \includegraphics[width=0.33\textwidth]{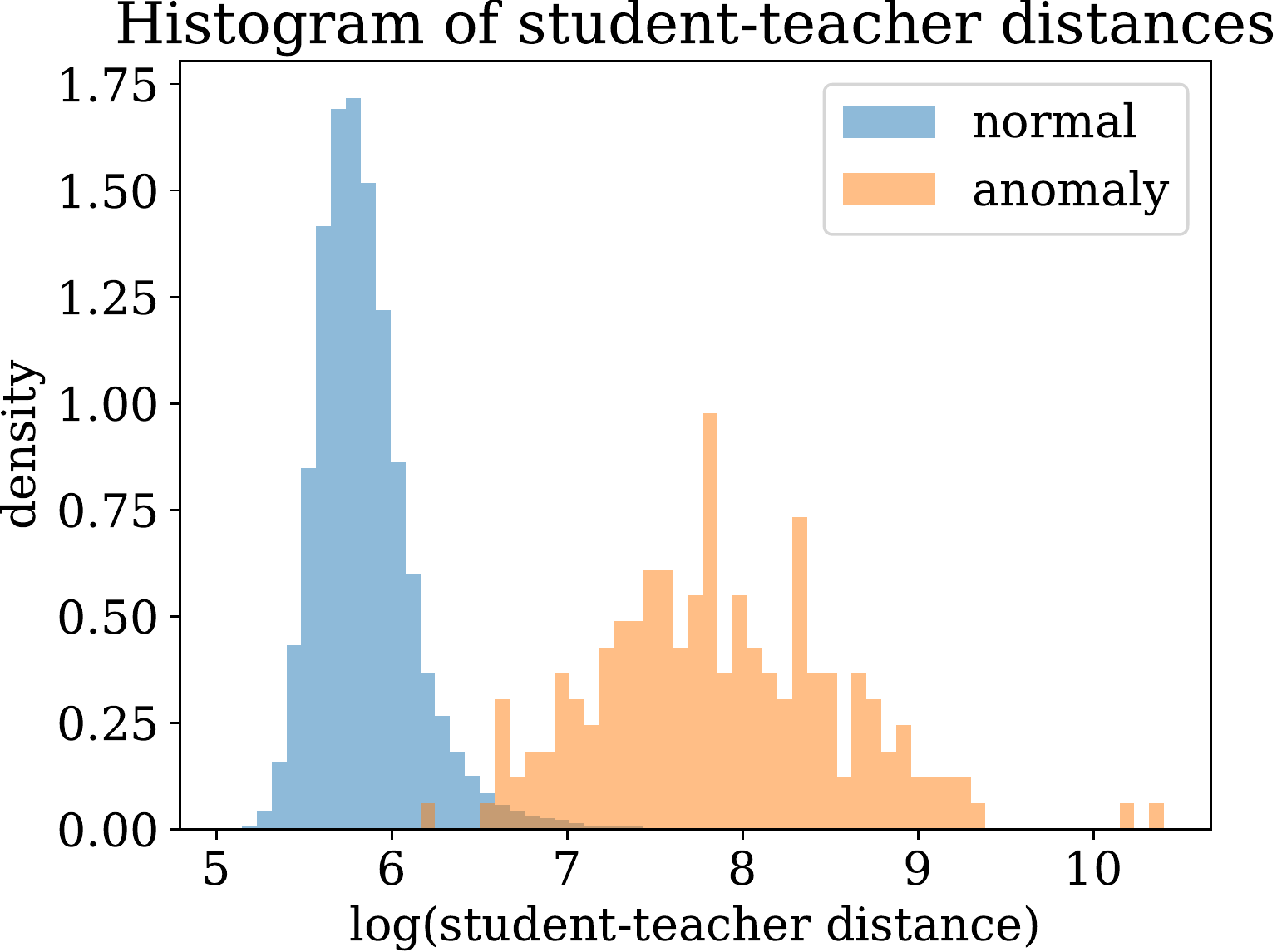} 
  \includegraphics[width=0.23\textwidth]{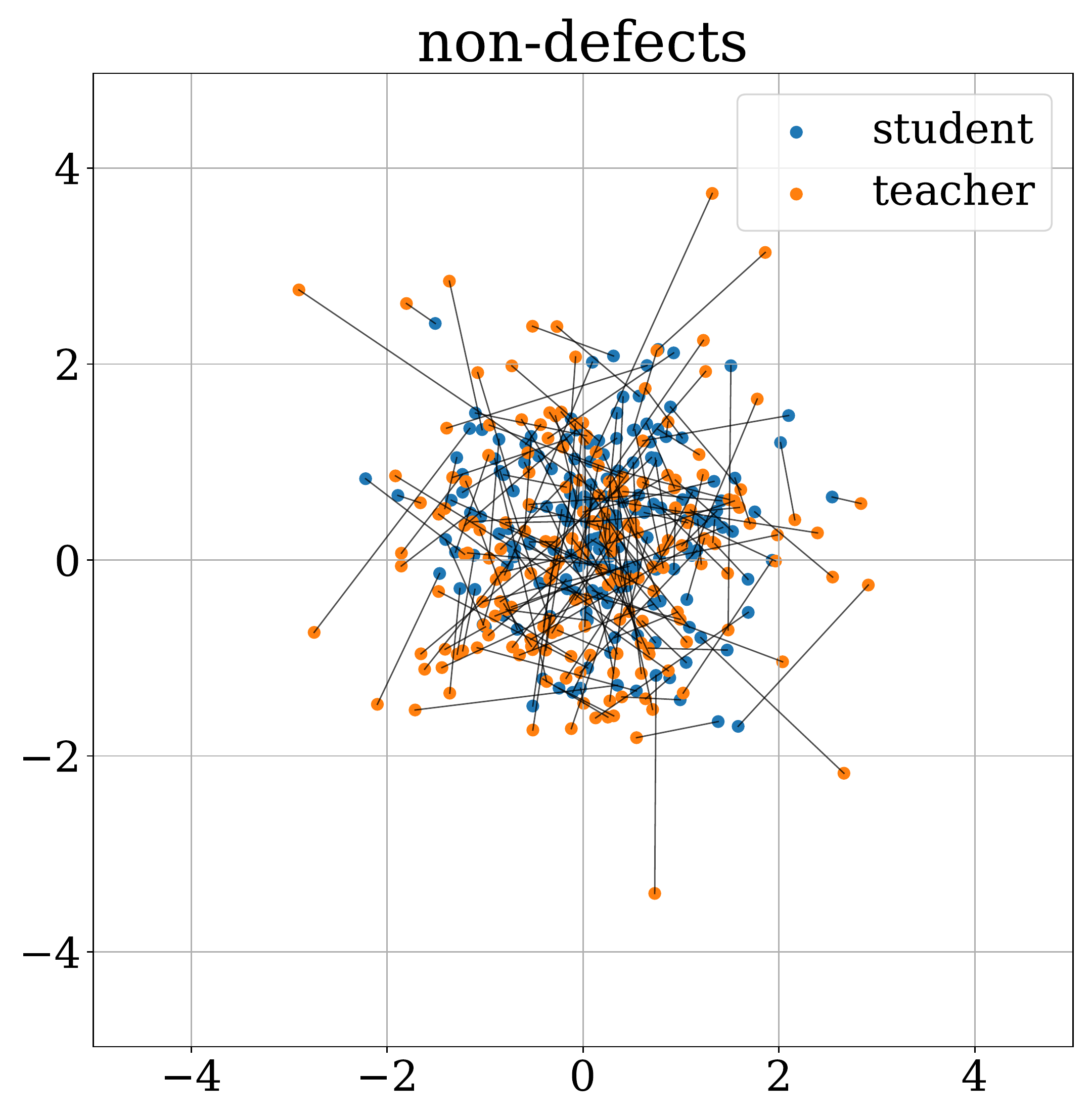} 
  \includegraphics[width=0.23\textwidth]{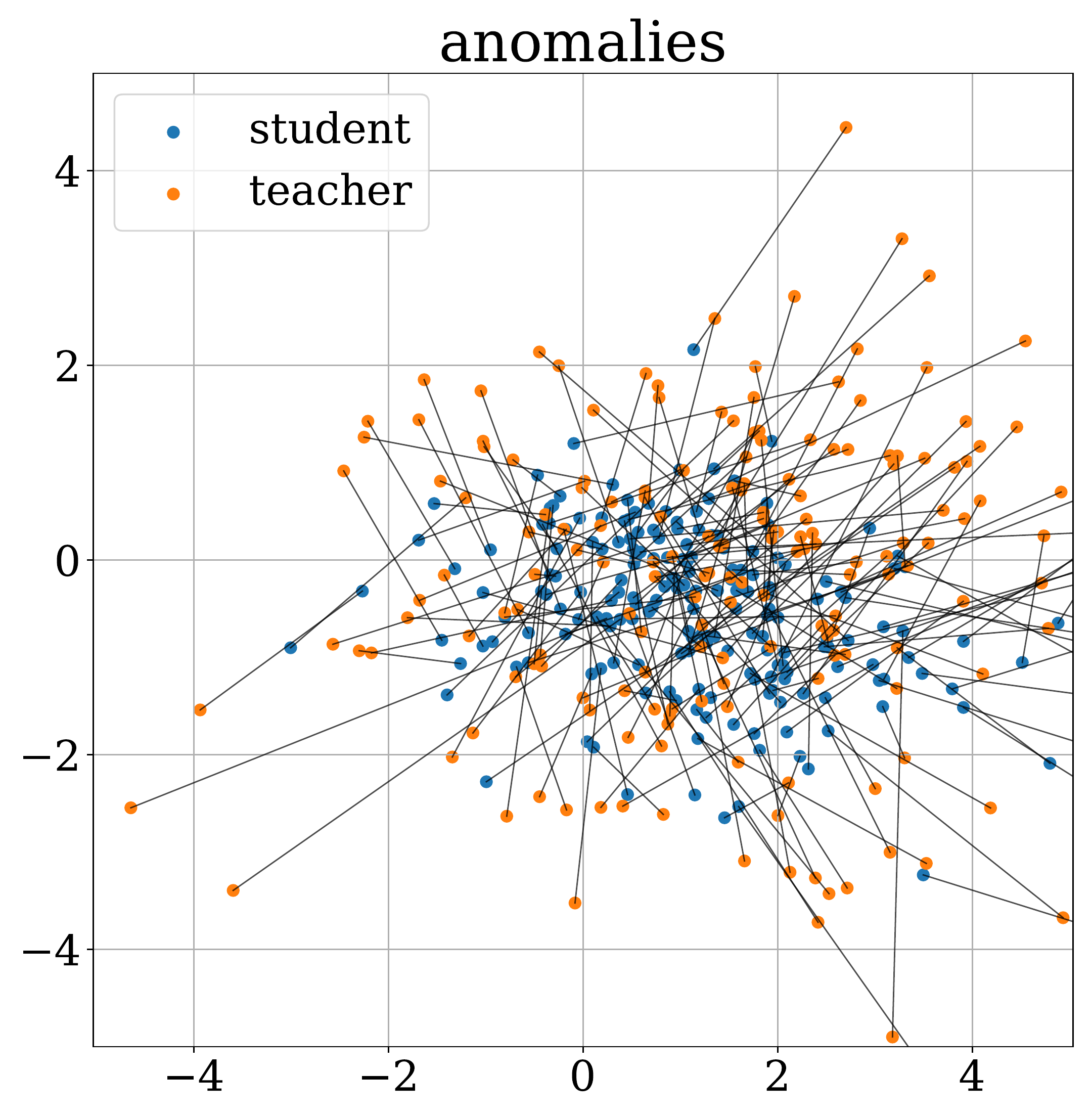}
 \caption{Top: Histogram of our AST distances for normal and anomalous regions of the class peach in MVT3D. Bottom: Random orthographic projections of student and teacher outputs grouped in non-defective (left plot) anomalous regions (right plot)  for the class peach.
 The plotted student-teacher distance representing the anomaly score is clearly higher for anomalous regions since the student is not able to match the teacher outputs, as it was only trained on non-defective regions.}
 \vspace{-0.25cm}
\label{fig:viz2d}
\end{figure}

For the class peach in the RGB+3D setting, the top of Figure \ref{fig:viz2d} compares the distribution of student-teacher distances for anomalous and normal regions.
The distribution of anomalous samples shows a clear shift towards larger distances.
At the bottom of Figure \ref{fig:viz2d}, the outputs of student and teacher as well as our the distance of corresponding pairs representing our anomaly score are visualized by a random orthographic 2D projection.
Note that visualizations made by techniques such as t-SNE~\cite{tsne} or PCA~\cite{pca} are not meaningful here, since the teacher outputs (and therefore most of the student outputs) follow an isotropic standard normal distribution.
Therefore, different random projections barely differ qualitatively.

\label{detection}

\subsubsection{Ablation Studies}
\label{ablation}
We demonstrate the effectiveness of our contributions and design decisions with several ablation studies.
Table \ref{table:ablation} compares the performance of variants of students with the teacher, which can be used as a density estimator itself for anomaly detection by using its likelihoods, given by Eq.~\ref{eqn:change_of_variables}, as anomaly score.
In comparison, a symmetric student-teacher pair worsens the results by 1 to 2\%, excepting the RGB case.
However, the performance is already improved for RGB and 3D+RGB by creating the asymmetry with a deeper version of the student than the teacher by doubling the number of coupling blocks to 8.
This effect is further enhanced if the architecture of the NF-teacher is replaced by a conventional feedforward network as we suggest.
We also vary the depth of our student network and analyzed its relation to performance, model size and inference time in Table \ref{table:depth}.
With an increasing number of residual blocks $n_{\mathrm{st\_blocks}}$, we observe an increasing performance which is almost saturated after 4 blocks.
Since the remaining potential in detection performance is not in relation to the linearly increasing additional computational effort per block, we suggest to choose 4 blocks to have a good trade-off.

In Table \ref{table:pe_mask} we investigate the impact of the positional encoding and the foreground mask.
For MVT3D, positional encoding improves the detection by 1.4\% of our AST-pair when trained with 3D data as the only input.
Even though the effect is not present when combining both data modalities, we consider it generally reasonable to use the positional encoding, considering that the integration with just 32 additional channels does not significantly increase the computational effort.

Foreground extraction in order to mask the loss for training and anomaly score for testing is also highly effective.
Since the majority of the image area often consists of background, the teacher has to spend a large part of the distribution on the background.
Masking allows the teacher and student to focus on the essential structures.
Moreover, noisy background scores are eliminated.

\begin{table}
\begin{center}
\footnotesize
    \begin{tabular}{l|c|c|c}
Method & 3D & RGB & 3D+RGB \\
    \hline
    Teacher only & 82.2 & 69.8 & 90.9\\
    NF student (symm.) & 81.8 & 76.0 & 88.9\\
    NF student (deeper) & 81.8 & 76.7 & 92.7\\
    \textbf{AST} (ours) & \textbf{83.3} & \textbf{88.0} & \textbf{93.7}\\
    \end{tabular}
\end{center}
\caption{Comparison of average detection performance in AUROC percentage on MVT3D of teacher and student-teacher in a symmetric and asymmetric setting. Our proposed asymmetric student-teacher pair outperforms all baselines in all cases.}
    \label{table:ablation}
\end{table}
\begin{table}
\small
\begin{center}
\footnotesize
    \begin{tabular}{c|c|c|c}
$n_{\mathrm{st\_blocks}}$ &  AUROC $[\%] \uparrow$ & \#Params. [M] $\downarrow$ & inf. time [ms] $\downarrow$\\
    \hline
    1 & 92.8 & 26.0 & 3.4\\
    2 & 93.3 & 44.8 & 6.1\\
    4 & 93.7 & 82.6 & 10.4\\
    8 & 93.7 & 151.1 & 19.8\\
    12 & 93.8 & 233.6 & 29.4\\
    \hline
    teacher & 90.9 & 3.8 & 4.5 \\
    \end{tabular}
\end{center}
\caption{Tradeoff between performance and computational effort on 3D+RGB data of MVT3D. The inference time was measured with a \textit{NVIDIA RTX 1080 Ti}.}
    \label{table:depth}
\end{table}
\begin{table}
\footnotesize
\begin{center}\begin{tabularx}{0.95\linewidth}{ c *{2}{|YY} }
input &  pos. enc. & mask & teacher & \textbf{AST} \\ \hline
     & \xmark& \checkmark&78.4 & 81.9 \\
    3D & \checkmark& \xmark& 59.4 & 67.2 \\
    & \checkmark& \checkmark& 82.2 & \textbf{83.3} \\ \hline
    & \xmark& \xmark& 69.3 & 87.8\\
    RGB & \checkmark& \xmark& 69.8 & \textbf{88.0} \\
    & \checkmark& \checkmark& n. a. & n. a.\\ \hline
    & \xmark& \checkmark& 90.9 & \textbf{93.8}\\
    3D+RGB & \checkmark& \xmark& 66.2 & 84.0 \\
    & \checkmark& \checkmark& 90.9 & 93.7 \\ 
    \end{tabularx}
\end{center}
\caption{Impact of the positional encoding and the foreground mask on the detection performance of student and teacher on MVT3D. Numbers are given in AUROC percentage. Since masks are obtained from 3D data, there is no mask for RGB. 
}
    \label{table:pe_mask}
\end{table}
\section{Conclusion}
We discovered the generalization problem of previous student teacher pairs for AD and introduced an alternative student-teacher method that prevents this issue by using a highly different architecture for student and teacher.
We were able to compensate for skewed likelihoods of a normalizing flow-based teacher, which was used directly for detection in previous work, by the additional use of a student.
Future work could extend the approach to more data domains and improve the localization resolution.

\vspace{-0.1em}
\small{\paragraph{Acknowledgements.}
This work was supported by the Federal Ministry of Education and
Research (BMBF), Germany under the project LeibnizKILabor (grant no.
01DD20003), the Center for Digital Innovations (ZDIN) and the Deutsche Forschungsgemeinschaft  (DFG) under  Germany’s  Excellence  Strategy  within  the  Cluster of Excellence PhoenixD (EXC 2122).
\newpage
\clearpage
{\small
\bibliographystyle{ieee_fullname}
\bibliography{egbib}

\begin{thebibliography}{10}\itemsep=-1pt

\bibitem{ganomaly}
Samet Akcay, Amir Atapour-Abarghouei, and Toby~P. Breckon.
\newblock Ganomaly: Semi-supervised anomaly detection via adversarial training.
\newblock In {\em Computer Vision -- ACCV 2018}, pages 622--637, Cham, 2019.
  Springer International Publishing.

\bibitem{amer2012nearest}
Mennatallah Amer and Markus Goldstein.
\newblock Nearest-neighbor and clustering based anomaly detection algorithms
  for rapidminer.
\newblock In {\em Proc. of the 3rd RapidMiner Community Meeting and Conference
  (RCOMM 2012)}, pages 1--12, 2012.

\bibitem{inn}
Lynton Ardizzone, Jakob Kruse, Sebastian Wirkert, Daniel Rahner, Eric~W
  Pellegrini, Ralf~S Klessen, Lena Maier-Hein, Carsten Rother, and Ullrich
  K{\"o}the.
\newblock Analyzing inverse problems with invertible neural networks.
\newblock In {\em ICLR}, 2019.

\bibitem{cinn}
Lynton Ardizzone, Carsten L{\"u}th, Jakob Kruse, Carsten Rother, and Ullrich
  K{\"o}the.
\newblock Guided image generation with conditional invertible neural networks.
\newblock {\em arXiv preprint arXiv:1907.02392}, 2019.

\bibitem{st_bergmann2}
Paul Bergmann, Kilian Batzner, Michael Fauser, David Sattlegger, and Carsten
  Steger.
\newblock Beyond dents and scratches: Logical constraints in unsupervised
  anomaly detection and localization.
\newblock {\em Int. J. Comput. Vis.}, 130(4):947--969, 2022.

\bibitem{mvtec}
Paul Bergmann, Michael Fauser, David Sattlegger, and Carsten Steger.
\newblock Mvtec ad--a comprehensive real-world dataset for unsupervised anomaly
  detection.
\newblock In {\em Proceedings of the IEEE Conference on Computer Vision and
  Pattern Recognition}, pages 9592--9600, 2019.

\bibitem{st_bergmann1}
Paul Bergmann, Michael Fauser, David Sattlegger, and Carsten Steger.
\newblock Uninformed students: Student-teacher anomaly detection with
  discriminative latent embeddings.
\newblock In {\em Proceedings of the IEEE/CVF Conference on Computer Vision and
  Pattern Recognition}, pages 4183--4192, 2020.

\bibitem{mvtec3d}
Paul Bergmann, Xin Jin, David Sattlegger, and Carsten Steger.
\newblock The mvtec 3d-ad dataset for unsupervised 3d anomaly detection and
  localization.
\newblock {\em 17th International Conference on Computer Vision Theory and
  Applications}, 2022.

\bibitem{ae_ssim}
Paul Bergmann, Sindy L{\"o}we, Michael Fauser, David Sattlegger, and C. Steger.
\newblock Improving unsupervised defect segmentation by applying structural
  similarity to autoencoders.
\newblock In {\em VISIGRAPP}, 2019.

\bibitem{lof}
Markus~M Breunig, Hans-Peter Kriegel, Raymond~T Ng, and J{\"o}rg Sander.
\newblock Lof: identifying density-based local outliers.
\newblock In {\em Proceedings of the 2000 ACM SIGMOD international conference
  on Management of data}, pages 93--104, 2000.

\bibitem{ADGAN}
Haoqing Cheng, Heng Liu, Fei Gao, and Zhuo Chen.
\newblock Adgan: A scalable gan-based architecture for image anomaly detection.
\newblock In {\em 2020 IEEE 4th Information Technology, Networking, Electronic
  and Automation Control Conference (ITNEC)}, volume~1, pages 987--993. IEEE,
  2020.

\bibitem{padim}
Thomas Defard, Aleksandr Setkov, Angelique Loesch, and Romaric Audigier.
\newblock Padim: a patch distribution modeling framework for anomaly detection
  and localization.
\newblock In {\em pattern Recognition, ICPR International Workshops and
  Challenges}, 2021.

\bibitem{imagenet}
Jia Deng, Wei Dong, Richard Socher, Li-Jia Li, Kai Li, and Li Fei-Fei.
\newblock Imagenet: A large-scale hierarchical image database.
\newblock In {\em 2009 IEEE conference on computer vision and pattern
  recognition}, pages 248--255. Ieee, 2009.

\bibitem{nf_trajectory}
Madson~LD Dias, C{\'e}sar Lincoln~C Mattos, Ticiana~LC da Silva, Jos{\'e}
  Ant{\^o}nio~F de Macedo, and Wellington~CP Silva.
\newblock Anomaly detection in trajectory data with normalizing flows.
\newblock {\em arXiv preprint arXiv:2004.05958}, 2020.

\bibitem{realnvp}
Laurent Dinh, Jascha Sohl-Dickstein, and Samy Bengio.
\newblock Density estimation using real nvp.
\newblock {\em ICLR 2017}, 2016.

\bibitem{DocBra2015}
Alexander Dockhorn, Christian Braune, and Rudolf Kruse.
\newblock An alternating optimization approach based on hierarchical
  adaptations of dbscan.
\newblock In {\em 2015 IEEE Symposium Series on Computational Intelligence
  (SSCI)}, number~2, pages 749--755, 2015.

\bibitem{DocBra2016}
Alexander Dockhorn, Christian Braune, and Rudolf Kruse.
\newblock Variable density based clustering.
\newblock In {\em 2016 IEEE Symposium Series on Computational Intelligence
  (SSCI)}, pages 1--8, Dec. 2016.

\bibitem{itae}
Ye Fei, Chaoqin Huang, Cao Jinkun, Maosen Li, Ya Zhang, and Cewu Lu.
\newblock Attribute restoration framework for anomaly detection.
\newblock {\em IEEE Transactions on Multimedia}, 2020.

\bibitem{georgescu2021anomaly}
Mariana-Iuliana Georgescu, Antonio Barbalau, Radu~Tudor Ionescu, Fahad~Shahbaz
  Khan, Marius Popescu, and Mubarak Shah.
\newblock Anomaly detection in video via self-supervised and multi-task
  learning.
\newblock In {\em Proceedings of the IEEE/CVF Conference on Computer Vision and
  Pattern Recognition}, pages 12742--12752, 2021.

\bibitem{memae}
Dong Gong, Lingqiao Liu, Vuong Le, Budhaditya Saha, Moussa~Reda Mansour, Svetha
  Venkatesh, and Anton van~den Hengel.
\newblock Memorizing normality to detect anomaly: Memory-augmented deep
  autoencoder for unsupervised anomaly detection.
\newblock In {\em Proceedings of the IEEE International Conference on Computer
  Vision}, pages 1705--1714, 2019.

\bibitem{gan}
Ian Goodfellow, Jean Pouget-Abadie, Mehdi Mirza, Bing Xu, David Warde-Farley,
  Sherjil Ozair, Aaron Courville, and Yoshua Bengio.
\newblock Generative adversarial nets.
\newblock In {\em Advances in neural information processing systems}, pages
  2672--2680, 2014.

\bibitem{cflow}
Denis Gudovskiy, Shun Ishizaka, and Kazuki Kozuka.
\newblock Cflow-ad: Real-time unsupervised anomaly detection with localization
  via conditional normalizing flows.
\newblock In {\em Proceedings of the IEEE/CVF Winter Conference on Applications
  of Computer Vision}, pages 98--107, 2022.

\bibitem{hinton2015distilling}
Geoffrey Hinton, Oriol Vinyals, Jeff Dean, et~al.
\newblock Distilling the knowledge in a neural network.
\newblock {\em arXiv preprint arXiv:1503.02531}, 2(7), 2015.

\bibitem{fpfh}
Eliahu Horwitz and Yedid Hoshen.
\newblock An empirical investigation of 3d anomaly detection and segmentation.
\newblock {\em arXiv preprint arXiv:2203.05550}, 2022.

\bibitem{batchnorm}
Sergey Ioffe and Christian Szegedy.
\newblock Batch normalization: Accelerating deep network training by reducing
  internal covariate shift.
\newblock In {\em International conference on machine learning}, pages
  448--456. PMLR, 2015.

\bibitem{adam}
Diederik~P Kingma and Jimmy Ba.
\newblock Adam: A method for stochastic optimization.
\newblock In {\em International Conference on Learning Representations (ICLR)},
  2015.

\bibitem{vae}
Diederik~P. Kingma and Max Welling.
\newblock Auto-encoding variational bayes.
\newblock {\em CoRR}, abs/1312.6114, 2013.

\bibitem{le2021perfect}
Charline Le~Lan and Laurent Dinh.
\newblock Perfect density models cannot guarantee anomaly detection.
\newblock {\em Entropy}, 23(12):1690, 2021.

\bibitem{cutpaste}
Chun-Liang Li, Kihyuk Sohn, Jinsung Yoon, and Tomas Pfister.
\newblock Cutpaste: Self-supervised learning for anomaly detection and
  localization.
\newblock In {\em Proceedings of the IEEE/CVF Conference on Computer Vision and
  Pattern Recognition}, pages 9664--9674, 2021.

\bibitem{isoforest}
Fei~Tony Liu, Kai~Ming Ting, and Zhi-Hua Zhou.
\newblock Isolation forest.
\newblock In {\em 2008 eighth ieee international conference on data mining},
  pages 413--422. IEEE, 2008.

\bibitem{mirzadeh2020improved}
Seyed~Iman Mirzadeh, Mehrdad Farajtabar, Ang Li, Nir Levine, Akihiro Matsukawa,
  and Hassan Ghasemzadeh.
\newblock Improved knowledge distillation via teacher assistant.
\newblock In {\em Proceedings of the AAAI Conference on Artificial
  Intelligence}, volume~34, pages 5191--5198, 2020.

\bibitem{nazare}
Tiago Nazare, Rodrigo de Mello, and Moacir Ponti.
\newblock Are pre-trained cnns good feature extractors for anomaly detection in
  surveillance videos?
\newblock {\em arXiv preprint arXiv:1811.08495}, 2018.

\bibitem{pca}
Karl Pearson.
\newblock Liii. on lines and planes of closest fit to systems of points in
  space.
\newblock {\em The London, Edinburgh, and Dublin philosophical magazine and
  journal of science}, 2(11):559--572, 1901.

\bibitem{nf}
Danilo Rezende and Shakir Mohamed.
\newblock Variational inference with normalizing flows.
\newblock In {\em International Conference on Machine Learning}, pages
  1530--1538. PMLR, 2015.

\bibitem{rippel}
Oliver Rippel, Patrick Mertens, and Dorit Merhof.
\newblock Modeling the distribution of normal data in pre-trained deep features
  for anomaly detection.
\newblock {\em arXiv preprint arXiv:2005.14140}, 2020.

\bibitem{patchcore}
Karsten Roth, Latha Pemula, Joaquin Zepeda, Bernhard Sch{\"o}lkopf, Thomas
  Brox, and Peter Gehler.
\newblock Towards total recall in industrial anomaly detection.
\newblock In {\em Proceedings of the IEEE/CVF Conference on Computer Vision and
  Pattern Recognition}, pages 14318--14328, 2022.

\bibitem{sae}
Marco Rudolph, Bastian Wandt, and Bodo Rosenhahn.
\newblock Structuring autoencoders.
\newblock In {\em Proceedings of the IEEE International Conference on Computer
  Vision Workshops}, 2019.

\bibitem{differnet}
Marco Rudolph, Bastian Wandt, and Bodo Rosenhahn.
\newblock Same same but differnet: Semi-supervised defect detection with
  normalizing flows.
\newblock In {\em Proceedings of the IEEE/CVF Winter Conference on Applications
  of Computer Vision}, pages 1907--1916, 2021.

\bibitem{csflow}
Marco Rudolph, Tom Wehrbein, Bodo Rosenhahn, and Bastian Wandt.
\newblock Fully convolutional cross-scale-flows for image-based defect
  detection.
\newblock In {\em Proceedings of the IEEE/CVF Winter Conference on Applications
  of Computer Vision}, pages 1088--1097, 2022.

\bibitem{fpfh_orig}
Radu~Bogdan Rusu, Nico Blodow, and Michael Beetz.
\newblock Fast point feature histograms (fpfh) for 3d registration.
\newblock In {\em 2009 IEEE international conference on robotics and
  automation}, pages 3212--3217. IEEE, 2009.

\bibitem{nf_deep}
Artem Ryzhikov, Maxim Borisyak, Andrey Ustyuzhanin, and Denis Derkach.
\newblock Normalizing flows for deep anomaly detection.
\newblock {\em arXiv preprint arXiv:1912.09323}, 2019.

\bibitem{anogan}
Thomas Schlegl, Philipp Seeb{\"o}ck, Sebastian~M Waldstein, Georg Langs, and
  Ursula Schmidt-Erfurth.
\newblock f-anogan: Fast unsupervised anomaly detection with generative
  adversarial networks.
\newblock {\em Medical image analysis}, 54:30--44, 2019.

\bibitem{nsa}
Hannah~M Schl{\"u}ter, Jeremy Tan, Benjamin Hou, and Bernhard Kainz.
\newblock Self-supervised out-of-distribution detection and localization with
  natural synthetic anomalies (nsa).
\newblock {\em arXiv preprint arXiv:2109.15222}, 2021.

\bibitem{nf_time_series}
Maximilian Schmidt and Marko Simic.
\newblock Normalizing flows for novelty detection in industrial time series
  data.
\newblock {\em arXiv preprint arXiv:1906.06904}, 2019.

\bibitem{ocsvm}
Bernhard Sch{\"o}lkopf, Robert~C Williamson, Alex Smola, John Shawe-Taylor, and
  John Platt.
\newblock Support vector method for novelty detection.
\newblock {\em Advances in neural information processing systems}, 12, 1999.

\bibitem{anoseg}
Jouwon Song, Kyeongbo Kong, Ye-In Park, Seong-Gyun Kim, and Suk-Ju Kang.
\newblock Anoseg: Anomaly segmentation network using self-supervised learning.
\newblock {\em arXiv preprint arXiv:2110.03396}, 2021.

\bibitem{efficientnet}
Mingxing Tan and Quoc Le.
\newblock Efficientnet: Rethinking model scaling for convolutional neural
  networks.
\newblock In {\em International Conference on Machine Learning}, pages
  6105--6114. PMLR, 2019.

\bibitem{tian2019contrastive}
Yonglong Tian, Dilip Krishnan, and Phillip Isola.
\newblock Contrastive representation distillation.
\newblock {\em arXiv preprint arXiv:1910.10699}, 2019.

\bibitem{tsne}
Laurens Van~der Maaten and Geoffrey Hinton.
\newblock Visualizing data using t-sne.
\newblock {\em Journal of machine learning research}, 9(11), 2008.

\bibitem{posenc}
Ashish Vaswani, Noam Shazeer, Niki Parmar, Jakob Uszkoreit, Llion Jones,
  Aidan~N Gomez, {\L}ukasz Kaiser, and Illia Polosukhin.
\newblock Attention is all you need.
\newblock {\em Advances in neural information processing systems}, 30, 2017.

\bibitem{wang2021student}
Guodong Wang, Shumin Han, Errui Ding, and Di Huang.
\newblock Student-teacher feature pyramid matching for anomaly detection.
\newblock {\em arXiv preprint arXiv:2103.04257}, 2021.

\bibitem{tomINN}
Tom Wehrbein, Marco Rudolph, Bodo Rosenhahn, and Bastian Wandt.
\newblock Probabilistic monocular 3d human pose estimation with normalizing
  flows.
\newblock {\em Proceedings of the IEEE International Conference on Computer
  Vision}, 2021.

\bibitem{xiao2021unsupervised}
Qinfeng Xiao, Jing Wang, Youfang Lin, Wenbo Gongsa, Ganghui Hu, Menggang Li,
  and Fang Wang.
\newblock Unsupervised anomaly detection with distillated teacher-student
  network ensemble.
\newblock {\em Entropy}, 23(2):201, 2021.

\bibitem{draem}
Vitjan Zavrtanik, Matej Kristan, and Danijel Sko{\v{c}}aj.
\newblock Draem-a discriminatively trained reconstruction embedding for surface
  anomaly detection.
\newblock In {\em Proceedings of the IEEE/CVF International Conference on
  Computer Vision}, pages 8330--8339, 2021.

\bibitem{dsebm}
Shuangfei Zhai, Yu Cheng, Weining Lu, and Zhongfei Zhang.
\newblock Deep structured energy based models for anomaly detection.
\newblock In {\em Proceedings of the 33rd International Conference on
  International Conference on Machine Learning-Volume 48}, pages 1100--1109,
  2016.

\bibitem{pixelunshuffle}
Kai Zhang, Wangmeng Zuo, and Lei Zhang.
\newblock Ffdnet: Toward a fast and flexible solution for cnn-based image
  denoising.
\newblock {\em IEEE Transactions on Image Processing}, 27(9):4608--4622, 2018.

\bibitem{adae}
Chong Zhou and Randy~C Paffenroth.
\newblock Anomaly detection with robust deep autoencoders.
\newblock In {\em Proceedings of the 23rd ACM SIGKDD international conference
  on knowledge discovery and data mining}, pages 665--674, 2017.

\end{thebibliography}
}

%








\end{document}